\documentclass[11pt]{article}

\usepackage[final]{acl}
\usepackage{booktabs}
\usepackage{times}
\usepackage{latexsym}
 \usepackage{multirow}
\usepackage[T1]{fontenc}

\usepackage[utf8]{inputenc}

\usepackage{microtype}

\usepackage{inconsolata}

\usepackage{graphicx}

\title{When Relations Break: Analyzing Relation Hallucination in Vision-Language Model Under Rotation and Noise}

\author{Philip Wootaek Shin, Ajay Narayanan Sridhar, Sivani Devarapalli,\\ \textbf{Rui Zhang, Jack Sampson, Vijaykrishnan Narayanan} \\
  The Pennsylvania State University \\
   \texttt{\{pws5345,afs6372,lfd5379,rmz5227,jms1257,vxn9\}@psu.edu}}

\begin{document}
\maketitle
\begin{abstract}
Vision–language models (VLMs) achieve strong multimodal performance but remain prone to relation hallucination, which requires accurate reasoning over inter-object interactions. We study the impact of visual perturbations, specifically rotation and noise, and show that even mild distortions significantly degrade relational reasoning across models and datasets. We further evaluate prompt-based augmentation and preprocessing strategies (orientation correction and denoising), finding that while they offer partial improvements, they do not fully resolve hallucinations. Our results reveal a gap between perceptual robustness and relational understanding, highlighting the need for more robust, geometry-aware VLMs.
\end{abstract}

\section{Introduction}

Vision–language models (VLMs) have been widely deployed across a range of applications, including object recognition~\cite{feng2025vision, jin2024llms}, scene understanding~\cite{liao2024vlm2scene, selvam2025simcache}, and multimodal reasoning~\cite{jiang2025vlm, xu2025visulogic}. Despite their impressive capabilities, these models are known to exhibit hallucinations, where the generated outputs are inconsistent with the visual input. Such hallucinations typically manifest in three forms: object hallucination, attribute hallucination, and relation hallucination~\cite{bai2025hallucinationmultimodallargelanguage}. Among these, relation hallucination remains particularly challenging, as it requires accurately capturing interactions and spatial relationships between objects rather than simply identifying their presence or attributes~\cite{zheng2025reefknotcomprehensivebenchmarkrelation}.

\begin{figure}[t]
\centering
\includegraphics[width=\linewidth]{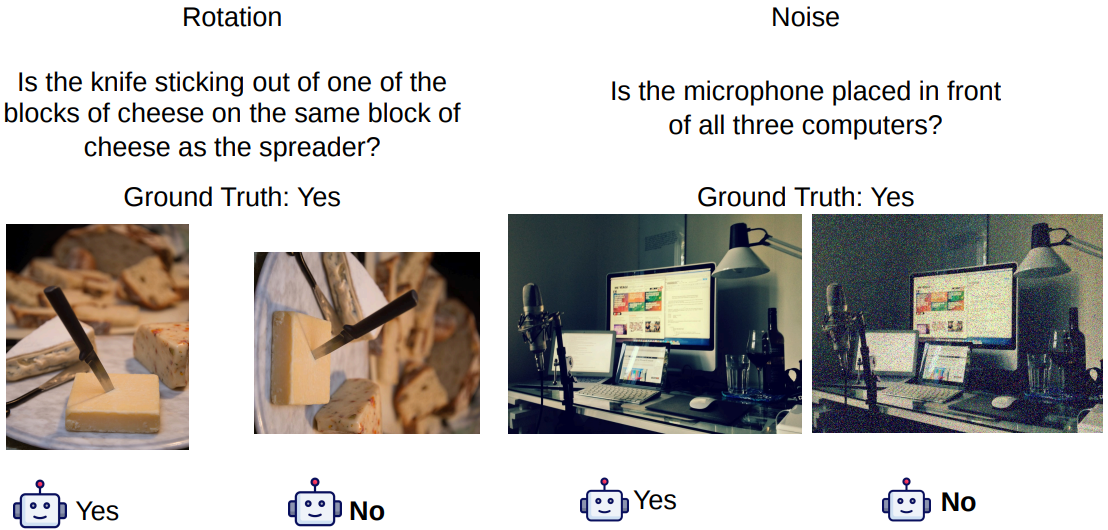}
\caption{VLM response under visual perturbations.}
\label{fig:task}
\end{figure}

In this work, we observe that relation hallucination is highly sensitive to visual perturbations, such as image noise and rotation, as illustrated in Fig.~\ref{fig:task}. While often studied independently, rotation and noise represent complementary failure modes—rotation disrupts geometric invariance, while noise degrades visual fidelity—and frequently co-occur in real-world settings, jointly affecting both structure and quality. While one might argue that rotating an image can alter perceived spatial relationships, humans naturally compensate for such transformations by mentally correcting orientation and still infer consistent object relations. In contrast, VLMs often fail to exhibit this invariance, leading to significantly degraded relational reasoning under such conditions.

To systematically investigate this phenomenon, we make the following contributions:
\begin{itemize}
\item Comprehensive Analysis: We conduct a systematic study of relation hallucination under varying  noise and rotation—capturing both geometric and photometric perturbations—across multiple VLMs and datasets, providing a unified evaluation of robustness.
\item Prompt-Based Intervention: We analyze how incorporating auxiliary prompts (e.g., rotation metadata or noise descriptions) influences model outcome and hallucination behavior.
\item Mitigation Strategy: We propose a practical mitigation approach by introducing preprocessing modules—such as rotation correction and denoising—prior to VLM inference, demonstrating improved robustness in relational reasoning.
\end{itemize}
\section{Related Work}

Hallucination in vision–language models can be categorized into object, attribute, and relation hallucination, with the latter being the most challenging due to its reliance on inter-object dependencies and spatial reasoning ~\cite{bai2025hallucinationmultimodallargelanguage}. Existing benchmarks study relation hallucination using yes/no and multiple-choice formats ~\cite{nie2025mmrelbenchmarkingrelationunderstanding, wu2024evaluatinganalyzingrelationshiphallucinations, zheng2025reefknotcomprehensivebenchmarkrelation}, but largely focus on clean visual inputs.

Recent work has explored VLM robustness under visual perturbations, showing that noise and rotation can significantly degrade performance~\cite{10.1109/TIFS.2024.3520306, shin2025losingplotvlmresponses, niu2026rotbenchevaluatingmultimodallarge}. However, these studies consider perturbations in isolation and do not directly address their impact on relation hallucination.

While preprocessing methods such as orientation detection and denoising ~\cite{barbosa2025deep_orientation, chen2025unirestore, yu2024scalingexcellencepracticingmodel} improve visual quality, their effectiveness in mitigating hallucination remains unclear. In contrast, our work systematically studies the combined effects of noise and rotation on relation hallucination and evaluates both prompt-based and preprocessing-based mitigation strategies, bridging robustness analysis with hallucination-specific evaluation.
\section{Rotation Analysis}

To evaluate our hypothesis on the effect of rotation on relation hallucination, we utilize three benchmark datasets: MMRel~\cite{nie2025mmrelbenchmarkingrelationunderstanding}, R-Bench~\cite{wu2024evaluatinganalyzingrelationshiphallucinations}, and Reefknot~\cite{zheng2025reefknotcomprehensivebenchmarkrelation}. Given that each dataset exhibits distinct characteristics and potential overlaps in image sources and annotations, we carefully curate subsets to ensure a fair and non-redundant evaluation. 

Specifically, we remove duplicate image–question pairs and avoid overlapping samples across datasets, while maintaining diversity in relational queries. Through this selective curation process, we construct evaluation sets comprising 1,632 image–question pairs for MMRel, 3,466 pairs for R-Bench, and Reefknot subsets with 1,185 multiple-choice questions and 2,922 binary (yes/no) questions. This rigorous selection enables a consistent and unbiased analysis of relation hallucination under rotational perturbations.

\subsection{Experimental Setup for Rotation-Induced Relation Hallucination}

To investigate the effect of image rotation on relation hallucination, we apply controlled clockwise (90$^\circ$) and counterclockwise (270$^\circ$) rotations to the input images. We evaluate both rotation directions using five open-source vision–language models—Qwen2-VL 7B~\cite{wang2024qwen2vl}, InternVL2 8B~\cite{chen2024internvl}, LLaVA-Next 8B~\cite{liu2024llavanext}, DeepSeek-Janus 7B~\cite{deepseek2024janus}, and LLaMA-3.2 11B (Vision)~\cite{meta2024llama32}—on the R-Bench dataset which we select as a representative benchmark due to its balanced coverage of relational reasoning tasks and manageable scale for controlled experimentation across multiple models. 

As shown in Tab.~\ref{tab:rotation_direction}, we observe no significant performance difference between 90 degrees clockwise and counterclockwise rotations. Therefore, for clarity and consistency, we report results using the clockwise rotation setting throughout the paper. We exclude 180° rotations, as such transformations are less representative of real-world viewing conditions and largely preserve global object configurations without introducing meaningful ambiguity in relational reasoning.

\begin{table}[t]
\centering
\footnotesize
\resizebox{\columnwidth}{!}{
\begin{tabular}{lccc}
\toprule
\textbf{Model} & \textbf{No Rotation (Orig)} & \textbf{90°} & \textbf{270°} \\
\midrule
Qwen2-VL 7B            & 78.02 & 72.07 & 72.65 \\
InternVL2 8B           & 79.80 & 75.36 & 75.62 \\
LLaVA-Next 8B          & 80.29 & 78.85 & 78.56 \\
DeepSeek-Janus 7B      & 64.66 & 64.66 & 64.66 \\
LLaMA-3.2 11B (Vision) & 81.16 & 77.29 & 77.44 \\
\bottomrule
\end{tabular}
}
\caption{Performance (\%) of open-source vision--language models on R-Bench under different rotation settings. 
}
\label{tab:rotation_direction}
\end{table}

For evaluation, we employ three widely used closed-source vision–language models: GPT-5.1~\cite{chatgpt5.1}, Gemini 2.5 Pro~\cite{gemini2.5pro}, and Claude Sonnet 4.5~\cite{claude-sonnet4.5}. These models are selected due to their strong multimodal reasoning capabilities and broad adoption in recent studies.

During inference, each model is provided with the rotated image and the original question prompt, without any additional guidance or metadata. The generated responses are then compared against the ground-truth answers provided in each dataset. Quantitative results and comparative analyses are summarized in Tab.~\ref{tab:rotation_results}. Overall, we observe a consistent performance degradation under clockwise rotation across all models and datasets, confirming the sensitivity of VLMs to geometric perturbations in relational reasoning task. We focus on closed-source VLMs as they represent state-of-the-art, widely deployed systems and must be evaluated under realistic black-box conditions, ensuring our findings generalize to real-world applications.

\begin{table}[t]
\centering
\footnotesize
\resizebox{\columnwidth}{!}{
\begin{tabular}{lccc}
\toprule
\textbf{Category} & \textbf{GPT-5.1} & \textbf{Gemini 2.5 Pro} & \textbf{Claude Sonnet 4.5} \\
\midrule
Reefknot Y/N (orig) & 76.61 & 75.41 & 72.25 \\
Reefknot Y/N (cw)   & 73.03 & 71.73 & 65.20 \\
\midrule
Reefknot MCQ (orig) & 88.51 & 86.08 & 80.08 \\
Reefknot MCQ (cw)   & 82.61 & 80.92 & 73.33 \\
\midrule
R-Bench Y/N (orig)  & 80.15 & 75.97 & 79.69 \\
R-Bench Y/N (cw)    & 78.51 & 48.93 & 72.74 \\
\midrule
MMRel Y/N (orig)    & 89.83 & 90.87 & 64.03 \\
MMRel Y/N (cw)      & 58.82 & 53.25 & 30.45 \\
\bottomrule
\end{tabular}
}
\caption{Accuracy (\%) of vision-language models under original (orig) and clockwise rotated (cw) settings across Reefknot, R-Bench, and MMRel datasets. 
}
\label{tab:rotation_results}
\end{table}

\subsection{Mitigating Rotation-Induced Relation Hallucination: Prompting vs. Preprocessing}

From Tab.~\ref{tab:rotation_direction} and Tab.~\ref{tab:rotation_results}, we observe that image rotation consistently increases relation hallucination, motivating the need for effective mitigation strategies. To address this issue, we explore two approaches: (1) prompt-based augmentation using rotation metadata and (2) preprocessing via image orientation correction.

\begin{figure}[t]
    \centering
    \includegraphics[width=\columnwidth]{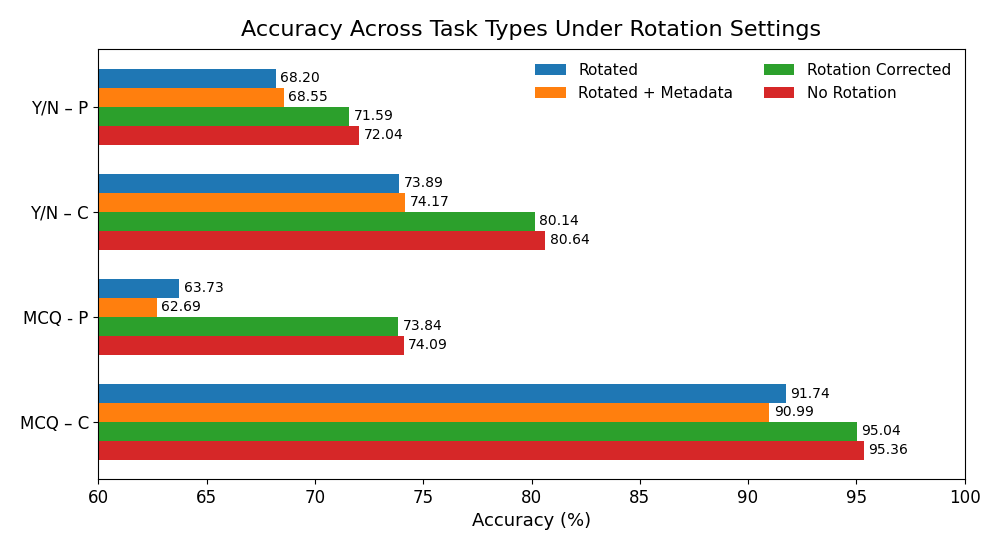}
    \caption{
    Effect of rotation metadata on VLM accuracy. (Reefknot, GPT-5.1) 
    }
    \label{fig:rotation_prompt}
\end{figure}

First, we investigate whether incorporating auxiliary prompts—specifically rotation information—can improve model robustness. Our hypothesis is that providing explicit orientation metadata may help VLMs compensate for geometric transformations and thus reduce hallucination. We conduct this analysis using GPT-5.1, which demonstrated relatively strong robustness on the Reefknot dataset under rotation. The Reefknot dataset contains both perceptual questions (orientation-related) and cognitive questions (action/interaction-related). However, as shown in Fig.~\ref{fig:rotation_prompt}, incorporating rotation metadata as an additional prompt yields minimal improvement in accuracy, suggesting that prompt-based guidance alone is insufficient to mitigate relation hallucination.

In contrast, we explore a preprocessing-based approach using a lightweight image orientation detector~\cite{barbosa2025deep_orientation}. We evaluate this detector on a modified Reefknot dataset with randomly applied 90° and 270° rotations. The detector achieves 99.66\% accuracy on multiple-choice questions (MCQ) with a downstream accuracy of 95.04\% (Cognitive) and 73.84\% (Perceptual) for GPT-5.1 and 99.38\% accuracy on yes/no (Y/N) questions with a downstream accuracy of 80.14\% (Cognitive) and 71.59\% (Perceptual) for GPT-5.1. These results indicate that correcting image orientation prior to VLM inference is a highly effective strategy, and is more reliable than relying on additional prompting to convey rotation information.


\section{Noise Analysis}
\noindent
To evaluate noise effects on relation hallucination, we adopt 19 corruption types from~\cite{10.1109/TIFS.2024.3520306, hendrycks2019benchmarking} across Reefknot, R-Bench, and MMRel, grouped into six categories (A–F). We exclude weather (E) and geometric/structural (F) corruptions, as they introduce unrealistic artifacts and alter scene geometry, thereby invalidating ground-truth object relationships.

From the remaining geometry-preserving categories (A–D: noise-based, blur/distortion, compression/resolution, and photometric/color corruptions), we select one representative corruption each—Impulse Noise, Gaussian Blur, Pixelate, and Saturate—to enable a controlled and interpretable evaluation while preserving spatial relationships critical for assessing relation hallucination.

\subsection{Impact of Noise Corruptions and Severity on Relation Hallucination}
\label{subsec:4.1}

To examine the effect of noise on relation hallucination, we apply four representative corruptions—Impulse Noise, Gaussian Blur, Pixelate, and Saturate—at severity level 2. We hypothesize that introducing such perturbations would increase hallucination by degrading visual fidelity and disrupting relational cues. We evaluate this using the Reefknot dataset with GPT-5.1. As shown in Tab.~\ref{tab:noise_results}, we observe a consistent decrease in accuracy across both yes/no and multiple-choice settings, as well as across perceptual and cognitive tasks, confirming the negative impact of noise on relational reasoning.

\begin{table}[t]
\centering
\footnotesize
\resizebox{\columnwidth}{!}{
\begin{tabular}{lccccc}
\toprule
\textbf{Category} & \textbf{Baseline} & \textbf{Gaussian Blur} & \textbf{Impulse Noise} & \textbf{Pixelate} & \textbf{Saturate} \\
\midrule
MC -- Perceptual  & 74.09\% & 62.18\% & 64.77\% & 63.47\% & 63.73\% \\
YN -- Perceptual  & 72.04\% & 69.69\% & 69.36\% & 69.88\% & 70.22\% \\
MC -- Cognitive    & 95.36\% & 91.99\% & 91.86\% & 92.61\% & 92.49\% \\
YN -- Cognitive  & 80.64\% & 75.05\% & 76.14\% & 76.92\% & 77.19\% \\
\bottomrule
\end{tabular}
}
\caption{Impact of noise corruptions (severity level 2) on relation hallucination performance using GPT-5.1 on the Reefknot dataset. Results are reported as correct predictions (accuracy \%). All corruption types lead to performance degradation compared to the baseline.}
\label{tab:noise_results}
\end{table}

\begin{figure}[t]
    \centering
    \includegraphics[width=\columnwidth]{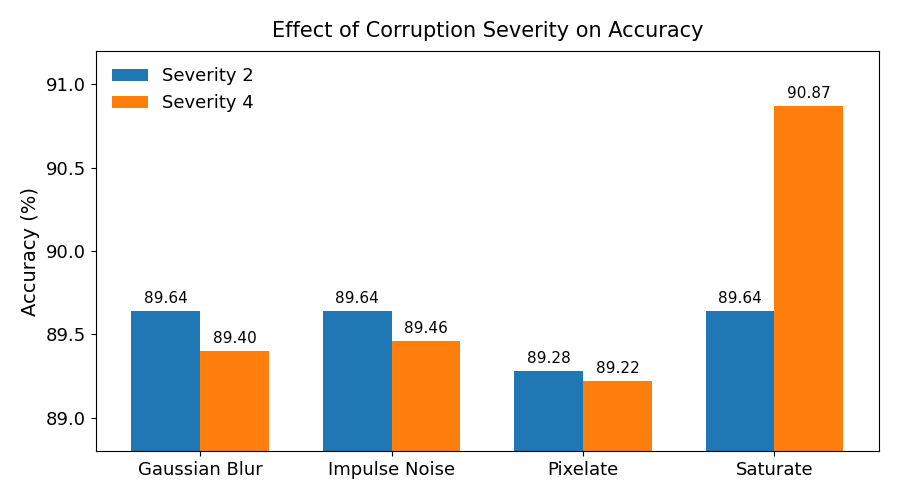}
    \caption{
    Effect of corruption severity on relation hallucination performance. (MMRel dataset, GPT-5.1) 
    }
    \label{fig:corruption}
\end{figure}

Furthermore, we analyze the effect of increasing corruption severity using GPT-5.1 on the MMRel dataset. As reported in Fig.~\ref{fig:corruption}, performance generally degrades as severity increases. However, \textit{saturate} occasionally improves accuracy, likely because enhanced color contrast strengthens salient cues, making object relationships easier for VLMs to detect.



\subsection{Mitigating Noise-Induced Relation Hallucination: Prompting vs. Denoising}

From Tab.~\ref{tab:noise_results} and Fig.~\ref{fig:corruption}, we observe that relation hallucination consistently worsens as image noise increases, confirming the sensitivity of VLMs to visual corruptions. Motivated by this, we explore two mitigation strategies: (1) prompt-based augmentation using noise information and (2) preprocessing via image denoising prior to model inference.

From Tab.~\ref{tab:combined_s2}, we compare the effectiveness of prompt-based augmentation and preprocessing-based denoising for mitigating noise-induced relation hallucination. Overall, both approaches provide limited but non-negligible improvements, with effectiveness varying across datasets and corruption types.
\begin{table}[h]
\centering
\footnotesize
\resizebox{\columnwidth}{!}{
\begin{tabular}{llcccc}
\toprule
\textbf{Dataset} & \textbf{Corruption} 
& \textbf{Acc (\%)} & \textbf{Drop (pp)} 
& \textbf{Acc-D (\%)} & \textbf{Drop-D (pp)} \\
\midrule
\multirow{4}{*}{MMRel}
 & Gaussian Blur & 86.70 & 3.12 & 89.34 & 0.49 \\
 & Impulse Noise & 87.32 & 2.51 & 88.97 & 0.86 \\
 & Pixelate      & 86.58 & 3.25 & 89.34 & 0.49 \\
 & Saturate      & 85.54 & 4.29 & 87.68 & 2.15 \\
\midrule
\multirow{4}{*}{R-Bench}
 & Gaussian Blur & 80.84 & 8.99 & 80.06 & 9.76 \\
 & Impulse Noise & 80.35 & 9.48 & 80.18 & 9.65 \\
 & Pixelate      & 80.24 & 9.59 & 80.18 & 9.65 \\
 & Saturate      & 78.04 & 11.78 & 77.44 & 12.39 \\
\bottomrule
\end{tabular}
}
\caption{Comparison of noise corruption (S2) and preprocessing-based denoising on relation hallucination across MMRel and R-Bench using GPT-5.1. Acc/Drop denote results under corruption, while Acc-D/Drop-D denote results after denoising.}
\label{tab:combined_s2}
\end{table}

On R-Bench, performance drops remain large (9–12 pp) under both settings, indicating that denoising alone cannot fully recover relational reasoning under severe noise. In contrast, on MMRel, denoising yields smaller drops (0.5–4 pp) and generally outperforms prompt-based augmentation, suggesting benefits when underlying visual structure is preserved.

For denoising, we evaluate state-of-the-art restoration models~\cite{chen2025unirestore, yu2024scalingexcellencepracticingmodel} and select outputs using LPIPS~\cite{zhang2018unreasonable}, PSNR, and SSIM~\cite{SSIM}. Despite improved perceptual quality, gains in relational reasoning are inconsistent, indicating a gap between low-level restoration and high-level understanding.

Overall, denoising is a useful but condition-dependent strategy: it helps when noise affects low-level features, but is less effective when relational cues are disrupted or task complexity is high

\section{Conclusion}

In this work, we study relation hallucination in VLMs under rotation and noise, showing that even simple perturbations significantly degrade relational reasoning, with larger effects on complex datasets. While prompting and preprocessing (orientation correction and denoising) provide partial improvements, they fail to fully recover performance, and gains in perceptual quality do not consistently translate to better reasoning. These results reveal a gap between perceptual robustness and relational understanding, motivating more robust, geometry-aware VLM designs.






\section*{Acknowledgments}

This work was supported in part by NSF Awards 2243979 and 2318101
 

\bibliography{custom}

@inproceedings{liao2024vlm2scene,
  title={VLM2Scene: Self-supervised image-text-LiDAR learning with foundation models for autonomous driving scene understanding},
  author={Liao, Guibiao and Li, Jiankun and Ye, Xiaoqing},
  booktitle={Proceedings of the AAAI Conference on Artificial Intelligence},
  volume={38},
  number={4},
  pages={3351--3359},
  year={2024}
}

@article{xu2025visulogic,
  title={Visulogic: A benchmark for evaluating visual reasoning in multi-modal large language models},
  author={Xu, Weiye and Wang, Jiahao and Wang, Weiyun and Chen, Zhe and Zhou, Wengang and Yang, Aijun and Lu, Lewei and Li, Houqiang and Wang, Xiaohua and Zhu, Xizhou and others},
  journal={arXiv preprint arXiv:2504.15279},
  year={2025}
}

@article{jiang2025vlm,
  title={VLM-R3: Region Recognition, Reasoning, and Refinement for Enhanced Multimodal Chain-of-Thought},
  author={Jiang, Chaoya and Heng, Yongrui and Ye, Wei and Yang, Han and Xu, Haiyang and Yan, Ming and Zhang, Ji and Huang, Fei and Zhang, Shikun},
  journal={arXiv preprint arXiv:2505.16192},
  year={2025}
}

@inproceedings{selvam2025simcache,
  title={SimCache: similarity caching for efficient VLM-based scene understanding},
  author={Selvam, Surya and Rajendran, Ravi K and Sankaradas, Murugan and Raghunathan, Anand and Chakradhar, Srimat T},
  booktitle={Proceedings of the Computer Vision and Pattern Recognition Conference},
  pages={3327--3336},
  year={2025}
}

@article{jin2024llms,
  title={Llms meet vlms: Boost open vocabulary object detection with fine-grained descriptors},
  author={Jin, Sheng and Jiang, Xueying and Huang, Jiaxing and Lu, Lewei and Lu, Shijian},
  journal={arXiv preprint arXiv:2402.04630},
  year={2024}
}

@article{feng2025vision,
  title={Vision-language model for object detection and segmentation: A review and evaluation},
  author={Feng, Yongchao and Liu, Yajie and Yang, Shuai and Cai, Wenrui and Zhang, Jinqing and Zhan, Qiqi and Huang, Ziyue and Yan, Hongxi and Wan, Qiao and Liu, Chenguang and others},
  journal={arXiv preprint arXiv:2504.09480},
  year={2025}
}

@article{hendrycks2019benchmarking,
  title={Benchmarking neural network robustness to common corruptions and perturbations},
  author={Hendrycks, Dan and Dietterich, Thomas},
  journal={arXiv preprint arXiv:1903.12261},
  year={2019}
}

@misc{bai2025hallucinationmultimodallargelanguage,
      title={Hallucination of Multimodal Large Language Models: A Survey}, 
      author={Zechen Bai and Pichao Wang and Tianjun Xiao and Tong He and Zongbo Han and Zheng Zhang and Mike Zheng Shou},
      year={2025},
      eprint={2404.18930},
      archivePrefixfx={arXiv},
      primaryClass={cs.CV},
      url={https://arxiv.org/abs/2404.18930}, 
}

@misc{nie2025mmrelbenchmarkingrelationunderstanding,
      title={MMRel: Benchmarking Relation Understanding in Multi-Modal Large Language Models}, 
      author={Jiahao Nie and Gongjie Zhang and Wenbin An and Yun Xing and Yap-Peng Tan and Alex C. Kot and Shijian Lu},
      year={2025},
      eprint={2406.09121},
      archivePrefix={arXiv},
      primaryClass={cs.CV},
      url={https://arxiv.org/abs/2406.09121}, 
}

@misc{wu2024evaluatinganalyzingrelationshiphallucinations,
      title={Evaluating and Analyzing Relationship Hallucinations in Large Vision-Language Models}, 
      author={Mingrui Wu and Jiayi Ji and Oucheng Huang and Jiale Li and Yuhang Wu and Xiaoshuai Sun and Rongrong Ji},
      year={2024},
      eprint={2406.16449},
      archivePrefix={arXiv},
      primaryClass={cs.CV},
      url={https://arxiv.org/abs/2406.16449}, 
}

@misc{zheng2025reefknotcomprehensivebenchmarkrelation,
      title={Reefknot: A Comprehensive Benchmark for Relation Hallucination Evaluation, Analysis and Mitigation in Multimodal Large Language Models}, 
      author={Kening Zheng and Junkai Chen and Yibo Yan and Xin Zou and Xuming Hu},
      year={2025},
      eprint={2408.09429},
      archivePrefix={arXiv},
      primaryClass={cs.LG},
      url={https://arxiv.org/abs/2408.09429}, 
}

@article{10.1109/TIFS.2024.3520306,
author = {Zhang, Hao and Shao, Wenqi and Liu, Hong and Ma, Yongqiang and Luo, Ping and Qiao, Yu and Zheng, Nanning and Zhang, Kaipeng},
title = {B-AVIBench: Toward Evaluating the Robustness of Large Vision-Language Model on Black-Box Adversarial Visual-Instructions},
year = {2025},
issue_date = {2025},
publisher = {IEEE Press},
volume = {20},
issn = {1556-6013},
url = {https://doi.org/10.1109/TIFS.2024.3520306},
doi = {10.1109/TIFS.2024.3520306},
journal = {Trans. Info. For. Sec.},
month = jan,
pages = {1434–1446},
numpages = {13}
}

@misc{niu2026rotbenchevaluatingmultimodallarge,
      title={RotBench: Evaluating Multimodal Large Language Models on Identifying Image Rotation}, 
      author={Tianyi Niu and Jaemin Cho and Elias Stengel-Eskin and Mohit Bansal},
      year={2026},
      eprint={2508.13968},
      archivePrefix={arXiv},
      primaryClass={cs.CV},
      url={https://arxiv.org/abs/2508.13968}, 
}

@misc{shin2025losingplotvlmresponses,
      title={Losing the Plot: How VLM responses degrade on imperfect charts}, 
      author={Philip Wootaek Shin and Jack Sampson and Vijaykrishnan Narayanan and Andres Marquez and Mahantesh Halappanavar},
      year={2025},
      eprint={2509.18425},
      archivePrefix={arXiv},
      primaryClass={cs.CV},
      url={https://arxiv.org/abs/2509.18425}, 
}

@misc{barbosa2025deep_orientation,
  author = {Duarte Barbosa},
  title = {Deep Image Orientation Detection},
  year = {2025},
  howpublished = {\url{https://github.com/duartebarbosadev/deep-image-orientation-detection}},
  note = {GitHub repository}
}

@inproceedings{chen2025unirestore,
  title={UniRestore: Unified Perceptual and Task-Oriented Image Restoration Model Using Diffusion Prior},
  author={Chen, I and Chen, Wei-Ting and Liu, Yu-Wei and Chiang, Yuan-Chun and Kuo, Sy-Yen and Yang, Ming-Hsuan and others},
  booktitle={Proceedings of the Computer Vision and Pattern Recognition Conference},
  pages={17969--17979},
  year={2025}
}

@misc{yu2024scalingexcellencepracticingmodel,
      title={Scaling Up to Excellence: Practicing Model Scaling for Photo-Realistic Image Restoration In the Wild}, 
      author={Fanghua Yu and Jinjin Gu and Zheyuan Li and Jinfan Hu and Xiangtao Kong and Xintao Wang and Jingwen He and Yu Qiao and Chao Dong},
      year={2024},
      eprint={2401.13627},
      archivePrefix={arXiv},
      primaryClass={cs.CV},
      url={https://arxiv.org/abs/2401.13627}, 
}

@article{SSIM,
  title={Image quality assessment: from error visibility to structural similarity},
  author={Wang, Zhou and Bovik, Alan C and Sheikh, Hamid R and Simoncelli, Eero P},
  journal={IEEE transactions on image processing},
  volume={13},
  number={4},
  pages={600--612},
  year={2004},
  publisher={IEEE}
}

@misc{gemini2.5pro,
  author = {Google DeepMind},
  title = {Gemini 2.5 Pro},
  year = {2026},
  howpublished = {\url{https://deepmind.google}},
  note = {Large language model}
}

@misc{claude-sonnet4.5,
  author = {Anthropic},
  title = {Claude Sonnet 4.5},
  year = {2026},
  howpublished = {\url{https://claude.ai}},
  note = {Large language model}
}

@misc{chatgpt5.1,
  author = {OpenAI},
  title = {ChatGPT-5.1},
  year = {2026},
  howpublished = {\url{https://chat.openai.com}},
  note = {Large language model}
}

@article{wang2024qwen2vl,
  title   = {Qwen2-VL: Enhancing Vision-Language Model's Perception of the World at Any Resolution},
  author  = {Wang, Peng and Bai, Shuai and Tan, Shengbang and Wang, Shuai and Fan, Zhihao and others},
  journal = {arXiv preprint arXiv:2409.12191},
  year    = {2024}
}

@article{chen2024internvl,
  title   = {InternVL: Scaling up Vision Foundation Models and Aligning for Generic Visual-Linguistic Tasks},
  author  = {Chen, Zhe and Wang, Wenhai and Cao, Yue and others},
  journal = {arXiv preprint arXiv:2312.14238},
  year    = {2024}
}

@article{liu2024llavanext,
  title   = {Improved Baselines with Visual Instruction Tuning},
  author  = {Liu, Haotian and Li, Chunyuan and Wu, Qingyang and Lee, Yong Jae},
  journal = {arXiv preprint arXiv:2310.03744},
  year    = {2024}
}

@article{deepseek2024janus,
  title   = {Janus: Decoupling Visual Encoding for Unified Multimodal Understanding and Generation},
  author  = {{DeepSeek-AI}},
  journal = {arXiv preprint arXiv:2410.13848},
  year    = {2024}
}

@article{meta2024llama32,
  title   = {Llama 3.2: Open Multimodal Foundation Models},
  author  = {{Meta AI}},
  journal = {arXiv preprint arXiv:2409.17146},
  year    = {2024}
}

@inproceedings{zhang2018unreasonable,
  title={The unreasonable effectiveness of deep features as a perceptual metric},
  author={Zhang, Richard and Isola, Phillip and Efros, Alexei A and Shechtman, Eli and Wang, Oliver},
  booktitle={Proceedings of the IEEE conference on computer vision and pattern recognition},
  pages={586--595},
  year={2018}
}




\end{document}